%% file: emnlp2021.tex
\pdfoutput=1
\documentclass[11pt]{article}

\usepackage{emnlp2021}

\usepackage{times}
\usepackage{latexsym}
\usepackage[smallerops]{newtxmath}
\usepackage{amsfonts}
\usepackage{xcolor}
\definecolor{mygreen}{RGB}{25, 150, 125}
\definecolor{myblue}{RGB}{12, 124, 186}
\definecolor{myred}{RGB}{201, 45, 57}
\usepackage[T1]{fontenc}
\usepackage[utf8]{inputenc}

\usepackage{microtype}
\usepackage{times}
\usepackage{latexsym}

\usepackage{amsmath}

\usepackage{amssymb}
\usepackage{booktabs}
\usepackage{enumitem}
\usepackage{algorithm}
\usepackage{algpseudocode}
\usepackage{multicol}
\usepackage{multirow}
\usepackage{xspace}
\usepackage{bm}
\usepackage{IEEEtrantools}
\usepackage{graphicx}

\newcommand{\mymodel}{LogiRE\xspace}

\title{ Learning Logic Rules for Document-level Relation Extraction}
\author{
\parbox{\linewidth}{
\centering
Dongyu Ru$^{\dagger\ddagger}$, 
Changzhi Sun$^{\ddagger}$\thanks{\ \ corresponding authors.} , Jiangtao Feng$^\ddagger$, 
Lin Qiu$^\dagger$,\\ 
Hao Zhou$^\ddagger$,
Weinan Zhang$^{\dagger}$\footnotemark[1] , 
Yong Yu$^\dagger$, 
Lei Li$^\mathsection$\thanks{\ \ Work is done while at ByteDance.}
}\\
$^\dagger$Shanghai Jiao Tong University ~~~~~~~~~~$^\ddagger$ByteDance AI Lab \\
$^\mathsection$University of California, Santa Barbara\\
{\texttt{\{maxru,lqiu,wnzhang,yyu\}@apex.sjtu.edu.cn}} \\
{\texttt{\{sunchangzhi,fengjiangtao,zhouhao.nlp\}@bytedance.com}}\\
{\texttt{lilei@cs.ucsb.edu}}
}

\begin{document}
\maketitle
\begin{abstract}
\input{sections/00abstract}
\end{abstract}

\section{Introduction}
\label{sec:intro}
\input{sections/01intro}

\section{Related Work}
\label{sec:related}
\input{sections/02related}

\section{Method}
\label{sec:method}
\input{sections/03method}
\section{Experiments}
\label{sec:exp}

\input{sections/04experiments_final}

\section{Conclusion}
\label{sec:conclusion}
\input{sections/05conclusion}

\section*{Acknowledgements}
The SJTU team is supported by Shanghai Municipal Science and Technology Major Project (2021SHZDZX0102) and National Natural Science Foundation of China (61772333). We thank Xinbo Zhang, Jingjing Xu, Yuxuan Song, Wenxian Shi and other anonymous reviewers for their insightful and detailed comments.

\bibliography{anthology,custom}
\bibliographystyle{acl_natbib}

\input{sections/06appendix}

\end{document}

%% file: sections/00abstract.tex
Document-level relation extraction aims to identify relations between entities in a whole document. 
Prior efforts to capture long-range dependencies have relied heavily on implicitly powerful representations learned through (graph) neural networks,
which makes the model less transparent.
%Recently, combining logic rules with neural networks has attracted more attention in NLP, such as knowledge graph reasoning.
To tackle this challenge,
in this paper, we propose \mymodel, a novel probabilistic model for document-level relation extraction by learning logic rules.
\mymodel treats logic rules as latent variables
and consists of two modules: a rule generator and a relation extractor.
The rule generator is to generate logic rules potentially contributing to final predictions, 
and the relation extractor outputs final predictions based on the generated logic rules.
Those two modules can be efficiently optimized with the expectation-maximization (EM) algorithm.
By introducing logic rules into neural networks, 
\mymodel can explicitly capture long-range dependencies as well as enjoy better interpretation.
%Empirical results show that \mymodel exhibits better performance over several strong baselines.
Empirical results show that \mymodel significantly outperforms several strong baselines in terms of relation performance ($\sim$1.8 F1 score) and logical consistency (over 3.3 logic score).
Our code is available at \url{https://github.com/rudongyu/LogiRE}.

%% file: sections/01intro.tex
%Relation extraction (RE) aims to identify the relations between entities in unstructured text, one of the core sub-tasks for information extraction. In the past years, detecting the semantic relations of entity pairs in a single sentence has been well explored and achieves great success. \cite{socher-etal-2012-semantic, dos-santos-etal-2015-classifying, han-etal-2018-hierarchical, zhang-etal-2018-graph}.
Extracting relations from a document has attracted significant research attention in information extraction (IE).
% Given a document with all entities,
% it aims to identify all semantic relations between given entities.
% One straightforward approach is to decompose the document level to the sentence level ~\cite{socher-etal-2012-semantic, zhang-etal-2018-graph}.
% However, it is impossible to extract relations across sentences,
% which is common in real-world scenarios.
Recently, instead of focusing on sentence-level ~\cite{socher-etal-2012-semantic, dos-santos-etal-2015-classifying, han-etal-2018-hierarchical, zhang-etal-2018-graph, wang-etal-2021-enpar, wang-etal-2021-unire},
researchers have turned to modeling directly at the document level ~\cite{wang2019fine, ye-etal-2020-coreferential, zhou2021atlop},
which provides longer context and requires more complex reasoning.
Early efforts focus mainly on learning a powerful relation (i.e., entity pair) representation, which implicitly captures long-range dependencies.
According to the input structure, we can divide the existing document-level relation extraction work into two categories: the \emph{sequence-based model}  and the \emph{graph-based model}.
\begin{figure}
    \centering
    \includegraphics[width=\linewidth]{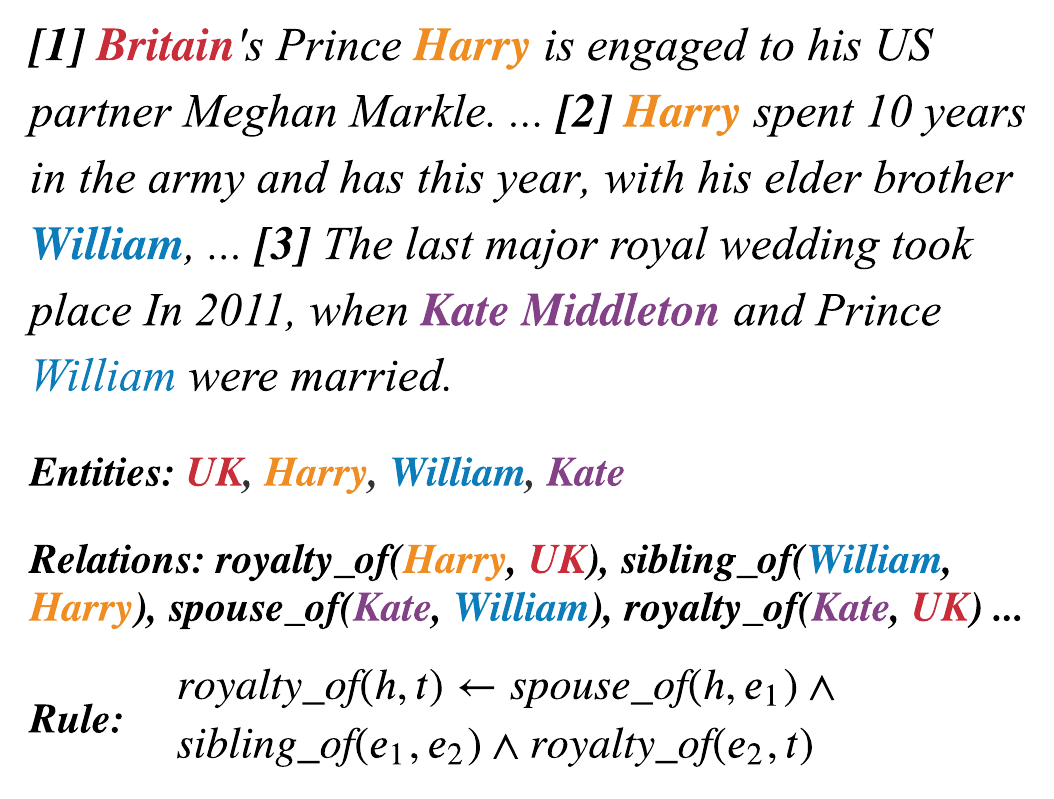}
    \caption{An example of relation identification by utilizing rules. The three labeled sentences describe the relations \textit{royalty\_of}(Harry,UK), \textit{sibling\_of}(William,Harry), and \textit{spouse\_of(Kate,William)}, respectively. The identification of the relation \textit{royalty\_of}(Kate,UK) requires the synthesis of information in three sentences. It can be easily derived from the demonstrated rule and the other three relations.}
    \label{fig:schematic}
    \vspace{-10pt}
\end{figure}

The sequence-based model first leverages different sequence encoder (e.g., BERT \cite{devlin-etal-2019-bert}, RoBERTa \cite{liu2019roberta}) to obtain token representations, 
and then computes relation representations by various pooling operations,
e.g., average pooling ~\cite{yao-etal-2019-docred, xu2021entity}, attentive pooling ~\cite{zhou2021atlop}.
%However, this sequence encoder plus pooling paradigm is difficult to capture interactions when the distance between words or entities is long.
%To alleviate this problem, many graph-based models are proposed.
To further capture long-range dependencies, graph-based models are proposed.
By constructing a graph, words or entities that are far away can become neighbor nodes.
On top of the sequence encoder, the graph encoder (e.g., GNN) can aggregate information from all neighbors, thus capturing longer dependencies.
Various forms of graphs are proposed, including
dependency tree ~\cite{peng-etal-2017-cross, zhang-etal-2018-graph},
co-reference graph ~\cite{sahu-etal-2019-inter},
mention-entity graph ~\cite{christopoulou-etal-2019-connecting, zeng-etal-2020-double},
entity-relation bipartite graph ~\cite{sun2019joint} and so on.
Despite their great success, there is still no comprehensive understanding of the internal representations, which are often criticized as mysterious "black boxes".

Learning logic rules can discover and represent knowledge in explicit symbolic structures that can be understood and examined by humans.
At the same time, logic rules provide another way to explicitly capture interactions between entities and output relations in a document.
For example in Fig.~\ref{fig:schematic}, the identification of \textit{royalty\_of(Kate,UK)} requires information in all three sentences. The demonstrated logic rule can be applied to directly obtain this relation from the three relations locally extracted in each sentence. Reasoning over rules bypasses the difficulty of capturing long-range dependencies and interprets the result with intrinsic correlations.
If the model could automatically learn rules and use them to make predictions, then we would get better relation extraction performance and enjoy more interpretation.

% In this paper, we propose \mymodel, a novel probabilistic model learning by logic rules.
In this paper, we propose \mymodel, a novel probabilistic model modeling intrinsic interactions among relations by logic rules.
Inspired by RNNLogic ~\cite{qu2020rnnlogic},
we treat logic rules as latent variables.
Specifically,
\mymodel consists of a rule generator and a relation extractor,
which are simultaneously trained to enhance each other.
The rule generator provides logic rules that are used by the relation extractor for prediction, 
and the relation extractor provides some supervision signals to guide the optimization of the rule generator, which significantly reduces the search space.
In addition, the proposed relation extractor is model agnostic, so it can be used as a plug-and-play technique for any existing relation extractors.
Those two modules can be efficiently optimized with the EM algorithm.
By introducing logic rules into neural networks, 
\mymodel can explicitly capture long-range dependencies between entities and output relations in a document and enjoy better interpretation.
Our main contributions are listed below:
\begin{itemize}[leftmargin=1pt, itemindent=1pc]
    \item We propose a novel probabilistic model for relation extraction by learning logic rules.
    The model can explicitly capture dependencies between entities and output relations, while enjoy better interpretation.
    \item We propose an efficient iterative-based method to optimize \mymodel based on the EM algorithm.
    \item Empirical results show that \mymodel significantly outperforms several strong baselines in terms of relation performance ($\sim$1.8 F1 score) and logical consistency (over 3.3 logic score).
\end{itemize}

%% file: sections/02related.tex
\begin{figure*}
    \centering
    \includegraphics[width=0.75\linewidth]{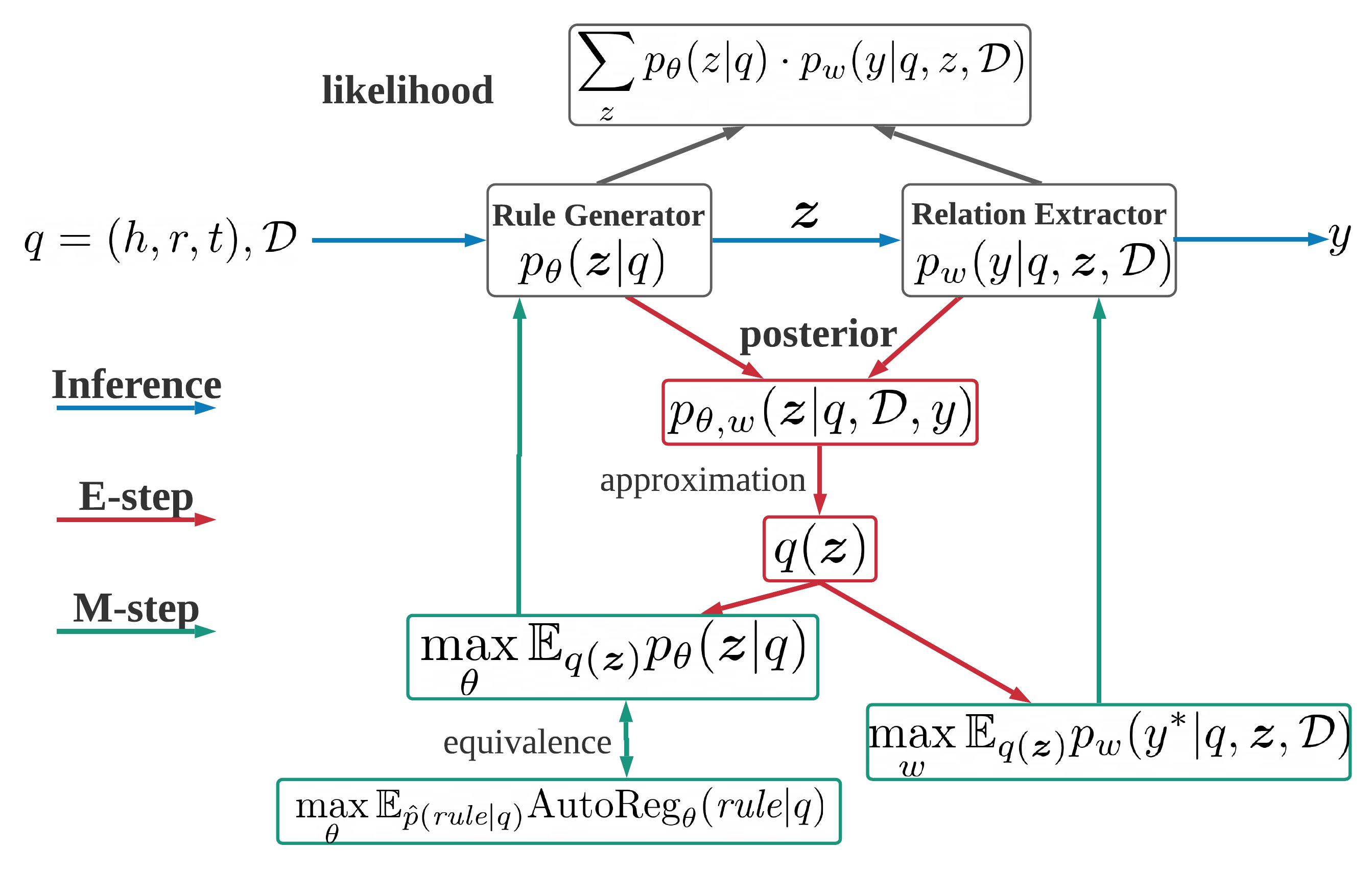}
    \caption{The overview of \mymodel. \mymodel consists of two modules: a rule generator $p_\theta$ and a relation extractor $p_w$. For a given document $\mathcal{D}$ and a query triple $q$, we treat the required logic rules as latent variables $\bm{z}$, aiming to identify the corresponding truth value $y$. During \textcolor{myblue}{inference}, we sample from the rule generator for the latent rule set and use the relation extractor to predict $y$ given the rules. The overall objective (maximizing the likelihood) is optimized by the EM algorithm. In the \textcolor{myred}{E-step}, we estimate the approximate posterior $q(\bm{z})$; In the \textcolor{mygreen}{M-step}, we maximize a lower bound of the likelihood w.r.t. $\theta, w$.}
    \label{fig:overview}
\end{figure*}

% Early work on relation extraction mainly focused on sentence-level relation extraction, aiming to classify the relation between a pair of entities within a sentence. Traditional approaches carefully design features\cite{kambhatla-2004-combining, zhou-etal-2005-exploring, jiang-zhai-2007-systematic, chan-roth-2010-exploiting} or kernels \cite{zelenko-etal-2002-kernel, culotta-sorensen-2004-dependency, mooney2006subsequence, zhang-etal-2006-composite, zhou-etal-2007-tree, qian-etal-2008-exploiting, nguyen-etal-2009-convolution, sun-han-2014-feature} for the classifier. Neural network-based methods refrained from the complicated feature engineering. They automatically learn the required features by leveraging various neural networks \cite{zeng-etal-2014-relation, zhang-etal-2015-bidirectional, cai-etal-2016-bidirectional, zhang-etal-2018-graph}. Many such approaches have been shown effective in capturing intra-sentence semantic relations. However, the identification of some relations may rely on inter-sentence correlations which they cannot handle.
%Document-level relation extraction has got more attention recently because of the richer information in the longer context and more complex interactions among entities and relations.
For document-level relation extraction,
prior efforts on capturing long-range dependencies mainly focused on two directions: pursuing stronger sequence representation \cite{nguyen-verspoor-2018-convolutional, verga-etal-2018-simultaneously, zheng2018effective} or including prior for interactions among entities as graphs \cite{christopoulou-etal-2019-connecting}.
For more powerful representations, they introduced pre-trained language models \cite{wang2019fine, ye-etal-2020-coreferential}, leveraged attentions for context pooling \cite{zhou2021atlop}, or integrated the scattered information according to a hierarchical level \cite{tang12084hin}.
Aiming to model the intrinsic interactions among entities and relations, they utilized implicit reasoning structures by carefully designing graphs connecting: mentions to entities, mentions in the same sentence \cite{christopoulou-etal-2019-connecting,sun2019joint}, mentions of the same entities \cite{wang-etal-2020-global, zeng-etal-2020-double}, etc. \citet{nan-etal-2020-reasoning, xu2021entity} directly integrated similar structural dependencies to attention mechanisms in the encoder.
These approaches contributed to obtaining powerful representations for distinguishing various relations but lacked interpretability on the implicit reasoning.
Another approach that can capture dependencies between relations is the global normalized model ~\cite{andor2016globally,sun2018extracting}.
In this work, we focus on how to learn and use logic rules to capture long-range dependencies between relations.

Another category of related work is logical reasoning.
Many studies were conducted on learning or applying logic rules for reasoning. Most of them \cite{qu2019probabilistic, zhang2020efficient} concentrated on reasoning over knowledge graphs, aiming to deduct new knowledge from existing triples. Neural symbolic systems \cite{hu-etal-2016-harnessing, wang-poon-2018-deep} combined logic rules and neural networks to benefit from regularization on deep learning approaches. These efforts demonstrated the effectiveness of integrating neural networks with logical reasoning. Despite doc-RE providing a suitable scenario for logical reasoning (with relations serving as predicates and entities as variables), no existing work attempted to learn and utilize rules in this field.
Using hand-crafted rules, \citet{wang2020integrating, wu-etal-2020-deep} achieved great success on sent-level information extraction tasks. However, the rules were predefined and limited to low-level operations, restricting their applications. 

%The optimization framework in our \mymodel is largely inspired by \citet{qu2020rnnlogic} to separately learn rules and rule weights for reasoning.

%% file: sections/03method.tex
%¥In this section, we illustrate our framework for both rule learning and rule utilization. In general, we assume the identification of relations between entities in long documents are dependent on potential rules. The rules are regarded as latent variables for prediction. We use a parameterized rule generator and a relation extractor to model the distribution of latent rules and the relation classification distribution given rules, respectively. We use an EM framework to jointly optimize the rule generator and the relation extractor. In Sec.~\ref{sec:method:pre}, we describe the task definition and the meaning of some terms and symbols. We picture the framework overview in Sec.~\ref{sec:method:overview} to present you the general idea. We show the process of rule learning and rule utilization in Sec.~\ref{sec:method:rule-learning} and Sec.~\ref{sec:method:rule-utilization}, respectively. The details on the optimization are illustrated in Sec.~\ref{sec:method:optimization}.

In this section, we describe the proposed method \mymodel that learns logic rules for document-level relation extraction.
We first define the task of document-level relation extraction and logic rules.
\paragraph{Document-level Relation Extraction}
Given a set of entities $\mathcal{E}$ with mentions scattered in a document $\mathcal{D}$, we aim to extract a set of relations $\mathcal{R}$.
A relation is a triple $(h, r, t) \in \mathcal{R}$ (also denoted by $r(h, t)$), where $h \in \mathcal{E}$ is the head entity, $t \in \mathcal{E}$ is the tail entity and $r$ is the relation type describing the semantic relation between two entities.
Let $\mathcal{T}_r$ be the set of possible relation types (including reverse relation types).
For simplicity, we define a query $q = (h, r, t)$ and
aim to model the probabilistic distribution $p(\bm{y}| q, \mathcal{D})$,
where $\bm{y} \in \{ -1, 1\}$ is a binary variable indicating whether $(h, r, t)$ is valid or not,
and $h, t \in \mathcal{E}, r \in \mathcal{T}_r$.
In this paper, bold letters indicate variables.

% \begin{figure*}
%     \centering
%     \includegraphics[width=5.0in]{figs/LogiRE.pdf}
%     \caption{overview}
%     \label{fig:overview}
%     \vspace{-10pt}
% \end{figure*}

\paragraph{Logic Rule}
We extract relations from the document by learning logic rules, where logic rules in this work have the conjunctive form:
\begin{IEEEeqnarray*}{c}
  \small
  \forall \{e_i \}_{i=0}^{l} ~ r(e_0, e_l) \leftarrow r_1(e_0, e_1) \land \dots \land r_l(e_{l-1}, e_l)
\end{IEEEeqnarray*}
 $e_i \in \mathcal{E}$, $r_i \in \mathcal{T}_r$ and $l$ is the rule length.
This form can express a wide range of common logical relations such as 
symmetry and transferability.
For example, transferability can be expressed as
\begin{IEEEeqnarray*}{c}
  \small
   \forall \{ e_0, e_1, e_2\} ~   r(e_0, e_2) \leftarrow r(e_0, e_1) \land r(e_1, e_2)
\end{IEEEeqnarray*}

Inspired by RNNLogic ~\cite{qu2020rnnlogic}, to infer high-quality logic rules in the large search space,
we separate rule learning and weight learning and treat the logic rules as the latent variable.
\mymodel consists of two main modules: the rule generator and the relation extractor, which are simultaneously trained to enhance each other.
Given the query $q = (h, r, t)$ in the document $\mathcal{D}$,
on the one hand,
the rule generator adopts an auto-regressive model to generate a set of logic rules based on $q$, which was used to help the relation extractor make the final decision;
on the other hand,
the relation extractor can provide some supervision signals to update the rule generator with posterior inference, 
which greatly reduces the search space with high-quality rules.

Unlike existing methods to capture the interactions among relations in the document by learning powerful representations,
we introduce a novel probabilistic model \mymodel (Sec.~\ref{sec:method:overview}, Fig.~\ref{fig:overview}),
which explicitly enhances the interaction by learning logic rules.
\mymodel uses neural networks to parameterize the rule generator and the relation extractor (Sec.~\ref{sec:method:parameterization}),
optimized by the EM algorithm in an iterative manner (Sec.~\ref{sec:method:optimization}).

\subsection{Overview}
\label{sec:method:overview}
We formulate the document-level relation extraction in a probabilistic way,
where a set of logic rules is assigned as a latent variable $\bm{z}$. 
Given a query variable $\bm{q} = (\bm{h}, \bm{r}, \bm{t})$ in the document $\mathcal{D}$,
we define the target distribution $p(\bm{y}| \bm{q}, \mathcal{D})$ as below\footnote{For simplicity, we omit $\mathcal{D}$ in distributions $p_{w, \theta}$ and $p_w$.}:
\begin{IEEEeqnarray*}{c}
  \small
  p_{w, \theta}(\bm{y}| \bm{q}) = \sum_{\bm{z}} p_w(\bm{y}|\bm{q}, \bm{z}) p_\theta(\bm{z} |\bm{q})
\end{IEEEeqnarray*}
where $p_\theta$ is the distribution of the rule generator which defines a prior over the latent variable $\bm{z}$ conditioned on a query $\bm{q}$ (we assume the distribution of $\bm {z}$ is independent from the document $\mathcal{D}$),
and $p_w$ is the relation extractor which gives the probability of $\bm{y}$ conditioned on the query $\bm{q}$, latent $\bm{z}$, and the document $\mathcal{D}$.
Given the gold label $y^*$ of the query $q$ in the document $\mathcal{D}$,
the objective function is to maximize the likelihood as follows:
\begin{IEEEeqnarray}{cl}
  \small
  \mathcal{L}(w, \theta) &= \log p_{w, \theta}(y^*| q) \label{eq:likelihood}
  %&= \log \mathbb{E}_{p_\theta(\bm{z}|q)} [p_w(y^*|q, \bm{z})] \nonumber
\end{IEEEeqnarray}
Due to the existence of latent variables in the objective function $\mathcal{L}$, we use the EM algorithm for optimization (Sec.~\ref{sec:method:optimization}).

\subsection{Parameterization}
\label{sec:method:parameterization}
We use neural networks to parameterize the rule generator and the relation extractor.
\paragraph{Rule Generator}
The rule generator defines the distribution $p_\theta(\bm{z}|\bm{q})$.
For a query ${q}$,
the rule generator generates a set of logic rules denoted by $\bm{z}$ for predicting the truth value $\bm y$ of the query ${q}$.

Formally, 
given a query ${q} = (h, r, t)$, 
we generate logic rules that takes the form of $r \leftarrow r_1 \land \dots \land r_l$.
Such relation sequences $[r_1, \dots, r_l]$ can be effectively modeled by an autoregressive model.
In this work, 
we employ a Transformer-based autoregressive model $\textrm{AutoReg}_\theta$ to parameterize the rule generator,
which sequentially generates each relation $r_i$.
In this process, the probabilities of generated rules are simultaneously computed.
%With such rule probabilities, we define the distribution over a set of rules $\bm{z}$ as a multinomial distribution:
Next, 
we assume that the rule set $\bm{z}$ obeys a multinomial distribution with $N$ rules independently sampled from the distribution $\textrm{AutoReg}_\theta(\emph{rule}|{q})$:
%$\emph{rule}\in\bm{z}$ denotes for a specific rule in $\bm{z}$, which determines the relation sequence in the rule body: $[r_1, \dots, r_l]$.
\begin{IEEEeqnarray*}{c}
   p_\theta(\bm{z}|{q}) \sim \textrm{Multi}(\bm{z} | N, \textrm{AutoReg}_\theta(\emph{rule}|{q})),
   \label{eq:multinomial}
\end{IEEEeqnarray*}
where $\textrm{Multi}$ denotes multinomial distribution, $N$ is a hyperparameter for the size of the set $\bm{z}$ and $\textrm{AutoReg}_\theta$ defines a distribution over logic rules conditioned on the query ${q}$.
\footnote{The generative process of a rule set $\bm{z}$ is quite intuitively similar to a translation model, and we simply generate $N$ rules with $\textrm{AutoReg}_\theta$ to form $\bm{z}$.}
%Generating a specific rule $\emph{rule}$ given the query $\bm{q}=(\bm{h}, \bm{r}, \bm{t})$ is regarded as a conditional sequence generation task in our method. We use a Transformer-based autoregressive model to generate the relation sequence in the rule body with the query triple provided as source. Intuitively, we translate the direct relation $\bm r$ between $\bm h$ and $\bm t$ to a specific meta path $[r_1, \dots, r_l]$ supporting $\bm r$.

{
%\color{blue}
\paragraph{Relation Extractor}
The relation extractor defines $p_w(\bm{y}| \bm{q}, \bm{z})$.  It utilizes a set of logic rules to get the truth value of $\bm{y}$ corresponding to the query $\bm{q}$. 
For each query ${q}$,
a $\emph{rule} \in \bm{z}$ is able to find different grounding paths on the document $\mathcal{D}$.
For example, Alice$\xrightarrow{\emph{father}}$Bob$\xrightarrow{\emph{spouse}}$Cristin is a grounding path for the rule $\emph{mother}(e_0, e_2)\leftarrow \emph{father}(e_0, e_1)\land\emph{spouse}(e_1, e_2)$.
Following the product t-norm fuzzy logic \cite{cignoli2000basic},
we score each \emph{rule}  as follows:
\begin{IEEEeqnarray*}{rl}
    \phi_w(\emph{rule}) &= \max_{\substack{\emph{path} \in \mathcal{P}(\emph{rule})}} \phi_w(\emph{path}) \\
        \emph{path} &:\mathop{{e}_0}_{({h})}\xrightarrow{r_1} {e}_1 \xrightarrow{r_2} {e}_2 \to \cdots \xrightarrow{r_l} \mathop{{e}_l}_{{(t)}} \\
        \phi_w(\emph{path}) &= \prod_{i=1}^l\phi_w({e}_{i-1}, r_i, {e}_i)
\end{IEEEeqnarray*}
where $\mathcal{P}(\emph{rule})$ is the set of grounding paths which start at ${h}$ and end at ${t}$ following a \emph{rule}.
$\phi_w({e}_{i-1}, {r}_i, {e}_i)$ is the confidence score obtained by any existing relation models. \footnote{This is why our approach is plug-and-play.}

%Following the inference of probabilistic logic programming \cite{de2015probabilistic,manhaeve2018deepproblog}, we evaluate the probability that the query $\bm{q}$ succeeds according to a bottom-up order: atoms, paths, rules, and then $p_w(\bm{y}| \bm{q}, \bm{z})$.

% The probability that each atom $r_i(e_{i-1}, e_i)$ succeeds is obtained by interpreting the confidence score $\phi_w(r_i(e_{i-1}, e_i)) \in [0, 1]$ from a backbone model as a probability.

% A path determined by each rule, namely a grounding of the rule body chain is evaluated by adopting the product t-norm fuzzy logic \cite{cignoli2000basic}:
% $\prod_{i=1}^l\phi_w(r_i(e_{i-1}, e_i))$.

% Given that a rule has multiple corresponding groundings in a document, we simply take the maximum evaluation result of all possible grounding paths as the probability for a rule. All grounding paths can be obtained by varying the intermediate entities $e_1,\cdots,e_{l-1}$. $\phi_w(\emph{rule},\mathcal{D})$ denotes for the probability that a rule's body succeeds in the document $\mathcal{D}$:
% \begin{IEEEeqnarray*}{c}
%     \emph{rule} \text{ body chain}:\bm{e}_0\xrightarrow{r_1} \bm{e}_1 \xrightarrow{r_2} \bm{e}_2 \to \cdots \xrightarrow{r_l} \bm{e}_l \\
%     \phi_w({\emph{rule}, \mathcal{D}}) = \max_{\substack{\bm{e}_1, \bm{e}_2,\dots,\bm{e}_{l-1}\in\mathcal{\mathcal{E}}\\\bm{e}_0=\bm{h},\bm{e}_l=\bm{t}}}\prod_{i=1}^l\phi_w(r_i(\bm{e}_{i-1},\bm{e}_i))
% \end{IEEEeqnarray*}

To get the probability (fuzzy truth value) of $\bm{y}$,
we synthesize the evaluation result of each rule in the latent rule set $\bm{z}$. 
The satisfaction of any rule body will imply the truth of $\bm{y}$.
Accordingly, we take the disjunction of all rules in $\bm{z}$ as the target truth value.
Following the principled sigmoid-based fuzzy logic function for disjunction \cite{sourek2018lifted, wang2020integrating}, we define the fuzzy truth value as:

{
\small
\begin{IEEEeqnarray*}{rl}
p_w(\bm{y}| {q}, \bm{z}) &= \mathrm{Sigmoid}(\bm{y}\cdot\mathtt{score}_w(q, \bm{z})) \\
\mathtt{score}_w(q, \bm{z}) &=  
\phi_w({q}) + \sum_{\emph{rule}\in\bm{z}}\phi_w({q}, \emph{rule})\phi_w(\emph{rule})
\end{IEEEeqnarray*}
}

where $\phi_w({q})$ and $\phi_w({q}, \emph{rule})$ are learnable scalar weights.
$\phi_w({q})$ is a bias term for balancing the score of positive and negative cases.
$\phi_w({q}, \emph{rule})$ estimates the score, namely, the quality of a specific rule.
$\phi_w(\emph{rule})$ evaluates the accessibility from the head entity $h$ to the tail entity $ t$ through the meta path defined by $\emph{rule}$'s body.
Applying logic rules and reasoning over the rules enable the relation extractor to explicitly modeling the long-range dependencies as the interactions among entities and relations.

}

\subsection{Optimization}
\label{sec:method:optimization}
To optimize the likelihood $\mathcal{L}(w, \theta)$ (Eq.~\ref{eq:likelihood}),
we update the rule generator and the relation extractor alternately in an iterative manner, namely the EM algorithm.
The classic EM algorithm estimates the posterior of the latent variable $\bm{z}$ according to current parameters in the E-step; 
The parameters are updated in the M-step with $\bm{z}$ obeys the estimated posterior.
However, in our setting, it is difficult to compute the exact posterior $p(\bm{z}|\bm{y},\bm{q})$ due to the large space of $\bm{z}$.
To tackle this challenge,
we seek an approximate posterior $q(\bm{z})$ by a second-order Taylor expansion.
This modified version of posterior forms a lower bound on $\log p_{w, \theta}(\bm{y} | \bm{q})$,
since the difference between them is a KL
divergence and hence positive:
\begin{IEEEeqnarray*}{ll}
\small
&\overbrace{\mathop{\mathbb{E}}_{q(\bm{z})} \left [\log \frac{p_{w, \theta}(\bm{y}, \bm{z} | \bm{q})}{p_{w, \theta}(\bm{z}|\bm{q}, \bm{y})} \right]} ^{\log p_{w, \theta}(\bm{y} | \bm{q})} -  \overbrace{\mathop{\mathbb{E}}_{q(\bm{z})} \left[\log \frac{p_{w, \theta}(\bm{y}, \bm{z}| \bm{q})}{q(\bm{z})}\right]}^\text{lower bound} \\
&= \mathrm{KL}\left(q(\bm{z}) || p_{w, \theta}(\bm{z} | \bm{q}, \bm{y})\right) \geq 0
\end{IEEEeqnarray*}
Once we get $q(\bm{z})$, 
we can maximize this lower bound of $\log p_{w, \theta}(\bm{y} | \bm{q})$.

\paragraph{E-step}
Given the current parameters $\theta, w$,
E-step aims to compute the posterior of $\bm{z}$ according to the current parameters $\theta, w$.
However, the exact posterior $p_{w, \theta}(\bm{z} | \bm{q}, \bm{y})$ is nontrivial due to its intractable partition function (space of $\bm{z}$ is large).
In this work, we aim to seek an approximate posterior $q(\bm{z})$.

By approximating the likelihood with the second-order Taylor expansion, we can obtain a conjugate form of the posterior as a multinomial distribution.
The detailed derivation is listed in Appendix.~\ref{sec:appendix:approx}. 
Formally, we first define $H(\emph{rule})$ as the score function estimating the quality of each rule:
\begin{IEEEeqnarray*}{l}
H(\emph{rule}) = \log \mathrm{AutoReg}_\theta(\emph{rule} | \bm{q}) + \\
\frac{y^*}{2}\left(\frac{1}{N}\phi_w(q)+\phi_w(q,\emph{rule})\phi_w(\emph{rule})\right)
\end{IEEEeqnarray*}
Intuitively, $H(\emph{rule})$ evaluates rule quality in two factors.
One is based on the rule generator $p_\theta$,
which servers as the prior probability for each \emph{rule}.
The other is based on the relation extractor,
and it takes into account the contribution of the current \emph{rule} to the final correct answer $y^*$.
Next, we use $\hat{p}(\emph{rule}|q)$ to denote the posterior distribution of the \emph{rule} given the query $q$:
\begin{IEEEeqnarray*}{l}
\hat{p}(\emph{rule}|q)\propto \exp\left(H(\emph{rule})\right)
\end{IEEEeqnarray*}
Thus the approximate posterior also obeys a multinomial distribution.
\begin{IEEEeqnarray*}{l}
q(\bm{z}) \sim \text{Multi}\left(N, \hat{p}(\emph{rule}|q)\right)
\end{IEEEeqnarray*}

\begin{algorithm}[t]
  \caption{EM Optimization for $\mathcal{L}(w, \theta)$}\label{alg:em}
  \begin{algorithmic}[1]
    \While {not converge}
        \State For each instance, use the rule generator $p_\theta$ to generate a set of logic rules $\hat{\bm{z}}(|\hat{\bm{z}}| = N)$.
        % \State For each instance, select $K$ high-quality logic rules $\hat{\bm{z}}_I$ from $\hat{\bm{z}}$ according to posterior distribution.
        \State Calculate the rule score $H(\emph{rule})$ of each rule for approximating the posterior of $\emph{rule}$: $\hat{p}(\emph{rule}|\bm{q})$.
        \Comment{E-step}
        \State For each instance, update the rule generator $\text{AutoReg}_\theta$ based on the sampled rules from $\hat{p}(\emph{rule}|\bm{q})$.
        \State For each instance, update the relation extractor $p_w$ based on generated logic rules $\hat{\bm{z}}$ from the updated rule generator.
        \Comment{M-step}
    \EndWhile
  \end{algorithmic}
  \label{alg:optimization}
\end{algorithm}

\paragraph{M-step}
After obtaining the $q(\bm{z})$,
M-step is to maximize the lower bound $\log p_{w, \theta}(\bm{y} | \bm{q})$ with respect to both $w$ and $\theta$. 
Formally, given each data instance $(y^*, q, \mathcal{D})$ and the $q(\bm{z})$, 
the objective is to maximize
\begin{IEEEeqnarray*}{l}
\small
{
\mathcal{L}_\mathrm{lower} = \overbrace{\mathop{\mathbb{E}}_{q(\bm{z})}[\log p_\theta(\bm{z}|q)]}^{\mathcal{L}_\mathrm{G}}   +
\overbrace{\mathop{\mathbb{E}}_{q(\bm{z})}[\log p_{w, \theta}(y^* |\bm{z},q)]}^{\mathcal{L}_\mathrm{R}}} \\
\end{IEEEeqnarray*}
where $\mathcal{L}_\mathrm{G}, \mathcal{L}_\mathrm{R}$ are the objective of the rule generator and  the relation extractor, repectively.

For the objective $\mathcal{L}_G$, it can be further converted equally as
\begin{IEEEeqnarray*}{c}
\mathcal{L}_\mathrm{G} = \mathbb{E}_{\hat{p}(\emph{rule}|q)}[\text{AutoReg}_\theta(\emph{rule}|q)]
\end{IEEEeqnarray*}
%There is an expectation operation with respect to ${\hat{p}(\emph{rule}|q)}$.
To compute the expectation term of $\mathcal{L}_\mathrm{G}$
we sample from the current prior $p_\theta(\bm{z}|q)$ for a sample $\hat{z}$, and evaluate the score of each rule as $H(\emph{rule})$, normalized score over $H(\emph{rule})$ are regarded as the approximated $\hat{p}(\emph{rule}|q)$.
Then we use sampled rules to update the $\text{AutoReg}_\theta(\emph{rule}|q)$.
Intuitively, we update the rule generator $p_\theta(\bm{z}|q)$ to make it consistent with the high-quality rules identified by the approximated posterior.

For the objective $\mathcal{L}_R$, we update the relation extractor according to the logic rules sampled from the updated rule generator. 
The logic rules explicitly capturing more interactions between relations can be fused as input to the relation extractor,
which yields better empirical results and enjoys better interpretation.
Finally, we summarize the optimization procedure in Algorithm \ref{alg:optimization}.

%% file: sections/04experiments_final.tex
We conduct experiments on multi-relational document-level relation extraction datasets: DocRED \cite{yao-etal-2019-docred} and DWIE \cite{zaporojets2020dwie}. The statistics of the two datasets are listed in Table~\ref{tab:dataset}. Pre-processing details of DWIE are described in Appendix~\ref{sec:appendix:imp}.

\begin{table}[t]
    \small
    \centering
    \begin{tabular}{llllll}
    \toprule
    \multicolumn{2}{c}{\bf Dataset} & \bf \#Doc. & \bf \#Rel. & \bf \#Ent. & \bf \#Facts \\ \midrule
    \multirow{3}{*}{DWIE} & train & 602 & \multirow{3}{*}{65} & 16494 & 14410  \\
    ~ & dev & 98 & ~ & 2785 & 2624 \\
    ~ & test & 99 & ~ & 2623 & 2459 \\ \midrule
    \multirow{3}{*}{DocRED} & train & 3053 & \multirow{3}{*}{96} & 59493 & 38180  \\
    ~ & dev & 1000 & ~ & 19578 & 12323 \\
    ~ & test & 1000 & ~ & 19539 & - \\ \bottomrule
    \end{tabular}
    \caption{Statistics of Document-level RE Datasets}
    \label{tab:dataset}
\end{table} 

\paragraph{Evaluation}
Besides the commonly used \textbf{F1} metric for relation extraction, we also include other two metrics for comprehensive evaluation of the models: \textbf{ign F1}, \textbf{logic}. \textbf{ign F1} was proposed in \cite{yao-etal-2019-docred} for evaluation with triples appearing in the training set excluded. It avoids information leakage from the training set. We propose \textbf{logic} for evaluation of logical consistency among the prediction results. Specifically, we use the 41 pre-defined rules on the DWIE dataset to evaluate whether the predictions satisfy these gold rules. The rules have a similar form to logic rules defined in Sec.~\ref{sec:method}. We name the precision of these rules on predictions as \textbf{logic} score. Note that these rules are independent of the rule learning and utilization in Sec.~\ref{sec:method} but only used for \textbf{logic} evaluation.

\paragraph{Experimental Settings}
The rule generator in our experimental settings is implemented as a transformer with a two-layer encoder and a two-layer decoder, hidden size set to 256. We empirically find the tiny structure is enough for modeling the required rule set. We set the size of the latent rule set $N$ to 50. We limit the maximum length of logic rules to 3 in our setting.

\subsection{Baselines}
We compare our \mymodel with the following baselines on document-level RE. The baselines are also used as corresponding backbone models in our framework. \citet{yao-etal-2019-docred} proposed to apply four state-of-the-art sentence-level RE models to document-level relation extraction: \textbf{CNN}, \textbf{LSTM}, \textbf{BiLSTM}, and \textbf{Context-Aware}. \cite{zeng-etal-2020-double} proposed \textbf{GAIN} to leverage both mention-level graph and aggregated entity-level graph to simulate the inference process in document-level RE, using graph neural networks.
\citet{zhou2021atlop} proposed \textbf{ATLOP}, using adaptive thresholding to learn a better adjustable threshold and enhancing the representation of entity pairs with localized context pooling. The implementation details of the baselines are shown in Appendix~\ref{sec:appendix:imp}.

\subsection{Main Results}
\begin{table*}[t]
    \scriptsize
    \centering
    \begin{tabular}{lccccccc}
    \toprule
        \multirow{2}{*}{\bf Model} & \multicolumn{3}{c}{\bf Dev} & \multicolumn{3}{c}{\bf Test} \\ \cmidrule(lr){2-4} \cmidrule(lr){5-7}
        & \bf ign F1 & \bf F1  & \bf logic & \bf ign F1 & \bf F1 & \bf logic \\ \midrule
          CNN  & 37.65 & 47.73 & 51.70 & 34.65 & 46.14 & 54.69 \\
         CNN + \mymodel & {40.31}(\underline{+2.65}) & {50.04}(\underline{+2.71}) & {72.84}(\underline{+21.14}) & {39.21}(\underline{+4.65}) & {50.44}(\underline{+4.30}) & {73.47}(\underline{+18.78}) \\
          \midrule
        LSTM  & 40.86 & 51.77 & 65.64 & 40.81 & 52.60 & 61.64 \\
        LSTM + \mymodel & 42.79(\underline{+1.93}) & 53.60(\underline{+1.83}) & 69.74(\underline{+4.10}) & 43.82(\underline{+3.01}) & 55.03(\underline{+2.43}) & 71.27(\underline{+9.63}) \\
        \midrule
        BiLSTM  & 40.46 & 51.92 & 64.87 & 42.03 & 54.47 & 64.41 \\
        BiLSTM + \mymodel & 42.59(\underline{+2.13}) & 53.83(\underline{+1.91}) & 73.37(\underline{+8.50}) & 43.65(\underline{+1.62}) & 55.14(\underline{+0.67}) & 77.11(\underline{+12.70}) \\
        \midrule
        Context-Aware & 42.06 & 53.05 & 69.27 & 45.37 & 56.58 & 70.01 \\
        Context-Aware + \mymodel & 43.88(\underline{+1.82}) & 54.49(\underline{+1.44}) & 73.98(\underline{+4.71}) & 48.10(\underline{+2.73}) & 59.22(\underline{+2.64}) & 75.94(\underline{+5.93}) \\
        \midrule
        GAIN  & 58.63 & 62.55 & 78.30 & 62.37 & 67.57 & 86.19 \\ 
        GAIN + \mymodel & 60.12 (\underline{+1.49}) & 63.91(\underline{+1.36}) & \textbf{87.86} (\underline{+9.56}) & \textbf{64.43} (\underline{+2.06}) & 69.40(\underline{+1.83}) & \textbf{91.22}(\underline{+5.02}) \\
        \midrule
        ATLOP  & 59.03 & 64.82  & 81.98 & 62.09 & 69.94 & 82.76 \\ 
        ATLOP + \mymodel & \textbf{60.24}(\underline{+1.21}) & \textbf{66.76}(\underline{+1.94}) & 86.98(\underline{+5.00}) & 64.11(\underline{+2.02})  & \textbf{71.78}(\underline{+1.84}) & 86.07(\underline{+3.31}) \\
        \bottomrule
    \end{tabular}
    \caption{Main results on DWIE. (The underlined statistics pass a t-test for significance with $p$ value < 0.01.)}
    \label{tab:dwie}
\end{table*}

% \begin{table}[t]
%     \small
%     \centering
%     \begin{tabular}{lcc}
%     \toprule
%         \multirow{2}{*}{\bf Model} & \multicolumn{2}{c}{\bf Dev} \\ \cmidrule(lr){2-3}
%         ~ & \bf ign F1 & \bf F1 \\ \midrule
%         GAIN & 57.82 & 59.99 \\ 
%          GAIN + \mymodel & 58.47(+0.65) & 60.62(+0.63) \\ \midrule 
%         ATLOP & 59.02 & 60.89 \\ 
%         ATLOP + \mymodel & 59.17(+0.15) & 61.33(+0.44) \\ \bottomrule
%     \end{tabular}
%     \caption{Comparison on DocRED. The improvements are less significant with reasons analyzed in Sec.~\ref{sec:exp:analysis:docred}.}
%     \label{tab:docred}
% \end{table}

\begin{table}[t]
    \small
    \centering
    \begin{tabular}{lcc}
    \toprule
        \multirow{2}{*}{\bf Model} & \multicolumn{2}{c}{\bf Test} \\ \cmidrule(lr){2-3}
        ~ & \bf ign F1 & \bf F1 \\ \midrule
        GAIN & 57.93 & 60.07 \\ 
         GAIN + \mymodel & 58.62(+0.69) & 60.61(+0.54) \\ \midrule 
        ATLOP & 59.14 & 61.13 \\ 
        ATLOP + \mymodel & 59.48(+0.34) & 61.45(+0.32) \\ \bottomrule
    \end{tabular}
    \caption{Comparison on DocRED. The improvements are less significant with reasons analyzed in Sec.~\ref{sec:exp:analysis:docred}.}
    \label{tab:docred}
\end{table}

Our \mymodel outperforms the baselines on all of the three metrics. (We mainly analyze the results on DWIE with all three metrics can be evaluated. The results on DocRED are demonstrated in Table~\ref{tab:docred} and discussed in Sec.~\ref{sec:exp:analysis:docred}.)

Our \mymodel consistently outperforms various backbone models. 
It outperforms various baselines on the DWIE dataset as shown in Table~\ref{tab:dwie}. We achieve 2.02 test ign F1 and 1.84 test F1 improvements on the current SOTA, ATLOP. The compatibility between \mymodel and various backbone models shows the generalization ability of our \mymodel. 
The consistent improvements on both sequence-based and graph-based models empirically verified the benefits of explicitly injecting logic rules to document-level relation extraction.

The improvements on graph-based models indicate the effectiveness of modeling interactions among multiple relations and entities. Despite graph-based models provide graphs \cite{christopoulou-etal-2019-connecting, wang-etal-2020-global} consisting of connections among mentions, entities, and sentences, they seek more powerful representations which implicitly model the intrinsic connections. Our \mymodel instead builds explicit interactions among the entities and relations through the meta path determined by the rules. The improvements on the current SOTA for graph-based model empirically proved the superiority of such explicit modeling.

Our model achieves better logical consistency compared with the baselines. The results show that \mymodel achieves up to 18.78 enhancement on the \textbf{logic} metric. Even on the graph-based model, GAIN, we obtain a significant improvement of 5.03 on logical consistency. The improved \textbf{logic} score shows that the predictions of \mymodel are more consistent with the regular logic patterns in the data. These numbers are evidence of the strength of our iterative-based optimization approach by introducing logic rules as latent variables.

\subsection{Analysis \& Discussion}
We analyze the results on DocRED data and discuss the superiority of our \mymodel on capturing long-range dependencies and interpretability. The capability of capturing long-range dependencies is studied by inspecting the inference performance on entity pairs of various distances. The interpretability is verified by checking the logic rules learned by our rule generator and the case study on predictions. 

\paragraph{Analysis on DocRED Results}
\label{sec:exp:analysis:docred}
In comparison with the significant improvements on DWIE, the enhancement of \mymodel on DocRED is less significant. Our analysis shows that the reasons are relatively shorter dependencies in DocRED and the logical inconsistency caused by incomplete annotations.

\begin{figure}
    \centering
    \includegraphics[width=\linewidth]{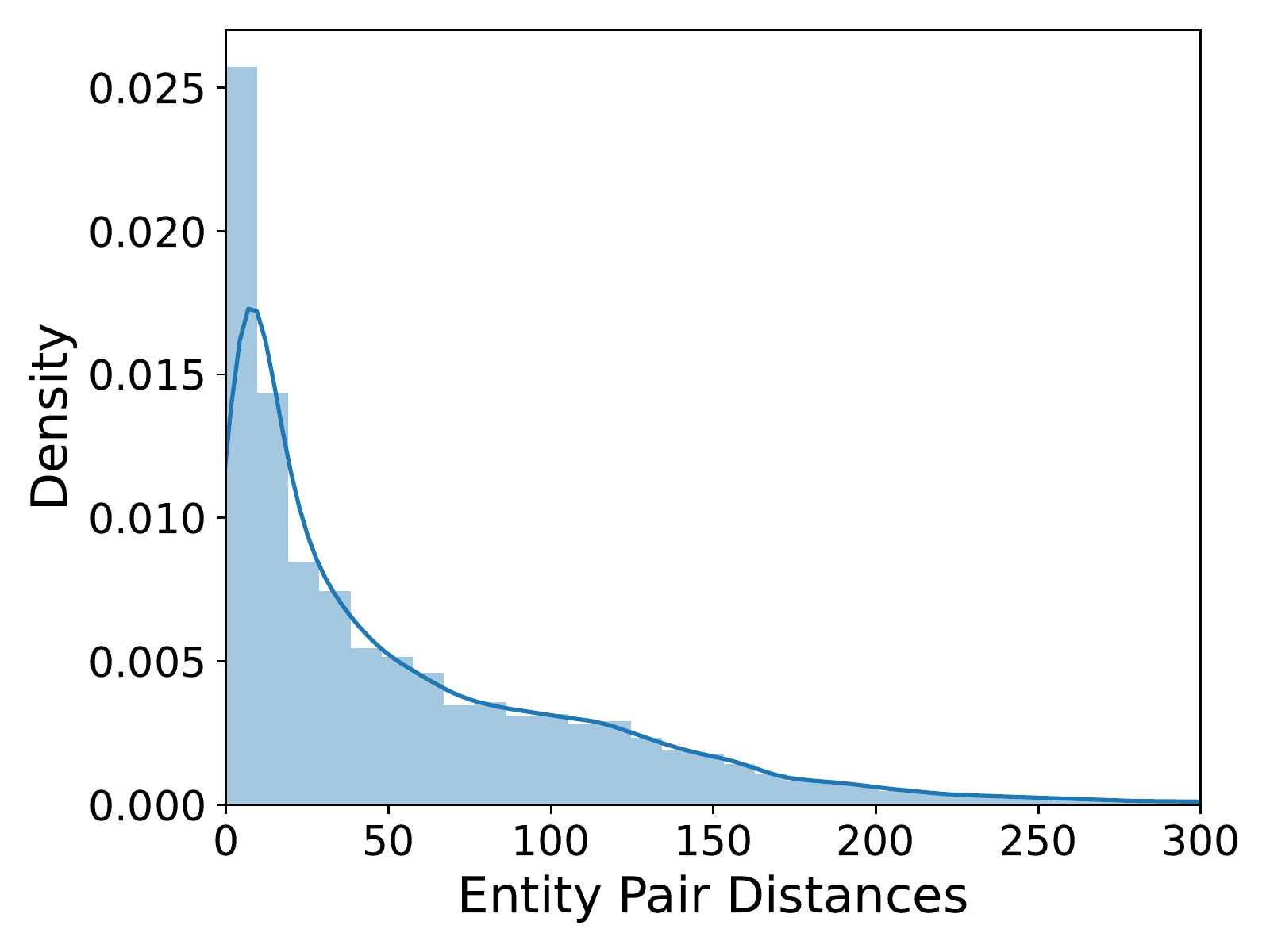}
    \caption{Distribution of Distance Between Entity Pairs in DocRED}
    \label{fig:dists-docred}
\end{figure}

\textbf{a) Shorter Dependencies in DocRED}
Shorter dependencies in DocRED lower the demand for capturing long-range correlations among entities and relations. We show the distribution of distance between entity pairs in Fig.~\ref{fig:dists-docred}. 79.26\% of entity pairs in DocRED have distances less than 100 tokens. The examples in DocRED are less difficult on capturing long-range dependencies. More analysis and comparison can be found in \citet{zaporojets2020dwie}. The representation-based approaches can already perform well in such cases. The benefits of modeling long-range dependencies through logical reasoning will be smaller.

\begin{table}[t]
    \centering
    \small
    \begin{tabular}{cc}
    \toprule
        implication rule & error rate \\ \midrule
        father$(h,z)\ \land$ spouse$(z, t)\rightarrow$ mother$(h, t)$ & 24.07\% \\
        replaces$^{-1}(h,t)\rightarrow$ replaced\_by$(h, t)$ & 22.22\% \\
        capital$^{-1}(h,t)\rightarrow$ capital\_of$(h, t)$ & 28.24\% \\
        father$^{-1}(h, t)\rightarrow$ child$(h, t)$ & 10.26\% \\
        followed$^{-1}(h, t)\rightarrow$ follows$(h, t)$ & 22.40\% \\
        capital$^{-1}(h, t)\rightarrow$ capital\_of$(h, t)$ & 28.24\% \\
        P150$^{-1}(h, t)\rightarrow$ P131$(h,t)$ & 19.71\% \\
    \bottomrule
    \end{tabular}
    \caption{The logical inconsistency in the DocRED (for conciseness, P150 represents the relation 'contains administrative territorial entity' and P131 represents the relation 'located in the administrative territorial entity'). The shown easy-to-verify gold rules have high error rates in DocRED while a considerable part of relations (12.96\%) are involved in as atoms in shown rules. Those missing annotations make the learning of logic rules difficult. Inconsistent patterns or statistics between training and test may lead to unfair evaluation of relation extraction performance.}
    \label{tab:docred-logic}
\end{table}

\textbf{b) Logical Inconsistency in DocRED}
The justification of predictions after reasoning may be not accurate because of missing annotations. We calculated the error rate of a few easy-to-verify logic rules as shown in Table.~\ref{tab:docred-logic}. The 7 rules, selected by case study, have a considerable part (12.96\%) of labeled relations may participate in as atoms. However, the statistics in the table demonstrated that all the 7 rules have error rates higher than 10\%. The numbers indicated that a notable partition of true relations are missing. The results obtained by reasoning over logic rules may be wrongly justified since the data is not exhaustively annotated.

According to the analysis above, our \mymodel has greater potential than that demonstrated as the overall performance on DocRED.

\begin{figure}
    \centering
    \includegraphics[width=\linewidth]{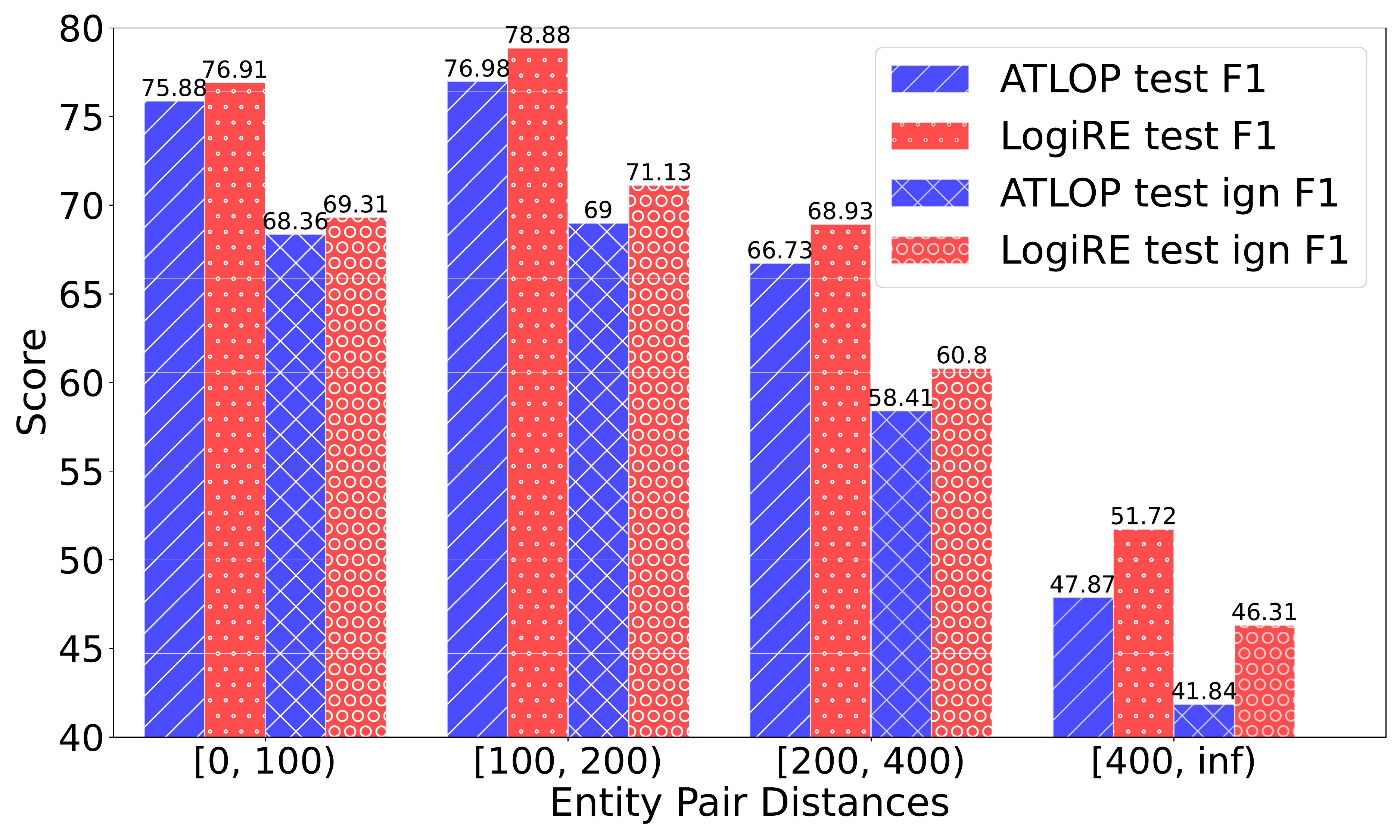}
    \caption{Performance gaps between ATLOP and \mymodel-ATLOP for entity pairs with different distances.}
    \label{fig:distance}
\end{figure}

\paragraph{Logic rules are shortcuts for comprehension.}
The performance enhancement of our \mymodel becomes more prominent when the distance between entity pairs gets longer. We plot the performance of ATLOP and ATLOP-based \mymodel on the DWIE dataset with four groups of entity pair distances in Fig.~\ref{fig:distance}. The distance is calculated as the number of tokens in between the nearest mentions of an entity pair. Results indicate that our \mymodel performs better on capturing long dependencies.

Relation extraction for entity pairs with longer distances in between generally performs worse. As shown in the figure, the performance starts to drop as the distance surpasses 100 tokens, indicating the difficulty of modeling long-range dependencies. The redundant information in a long context impedes accurate semantic mapping through powerful representations. This issue increases the complexity of modeling and limits the potential of representation-based approaches.

Our framework with latent logic rules injected can effectively alleviate this problem. The performance drop of our \mymodel is smaller when the distance between entities gets larger. For entity pairs of distances larger than 400, our \mymodel achieves up to 4.47 enhancement on test ign F1. By reasoning over local logic units (atoms in rules), we ignore the noisy background information in the text but directly integrate high-level connections among concepts to get the answer.

The reasoning process of our \mymodel is in line with the comprehension way of we human beings when reading long text. We construct basic concepts and connections between (local logic atoms) for each local part of the text. When the collected information is enough to fit some prior knowledge (logic rules), we deduct new cognition from the existing knowledge. Our \mymodel provides shortcuts for modeling long text semantics by adding logic reasoning to naive semantics mapping.

\begin{table}[t]
    \small
    \centering
    \begin{tabular}{c}
    \toprule
        played\_by$(h, z)\ \land\ $plays\_in$(z, t) \rightarrow$ character\_in $(h, t)$  \\ \hline
        (parent\_of$(h, z)\ \lor \ $child\_of$(h, z)\ \lor \ $spouse\_of$(h, z)$) \\ $\land$ royalty\_of$(z, t)\ \rightarrow$ royalty\_of$(h,t)$  \\ \hline
        event\_in2$^{-1}(h,z)\ \land$ event\_in0$(z, t)\rightarrow$ in0$(h, t)$ \\ \hline
        minister\_of$(h, z)\ \land$ in0$(z, t)\rightarrow$ citizen\_of$(h, t)$ \\ \hline
        member\_of$^{-1}(h, z)\ \land$ agent\_of$(z, t)\rightarrow$ based\_in0$(h, t)$\\
        \bottomrule
    \end{tabular}
    \caption{Example rules extracted from \mymodel trained on the DWIE dataset.}
    \label{tab:induct}
\end{table}

\begin{figure*}
    \centering
    \includegraphics[width=\linewidth]{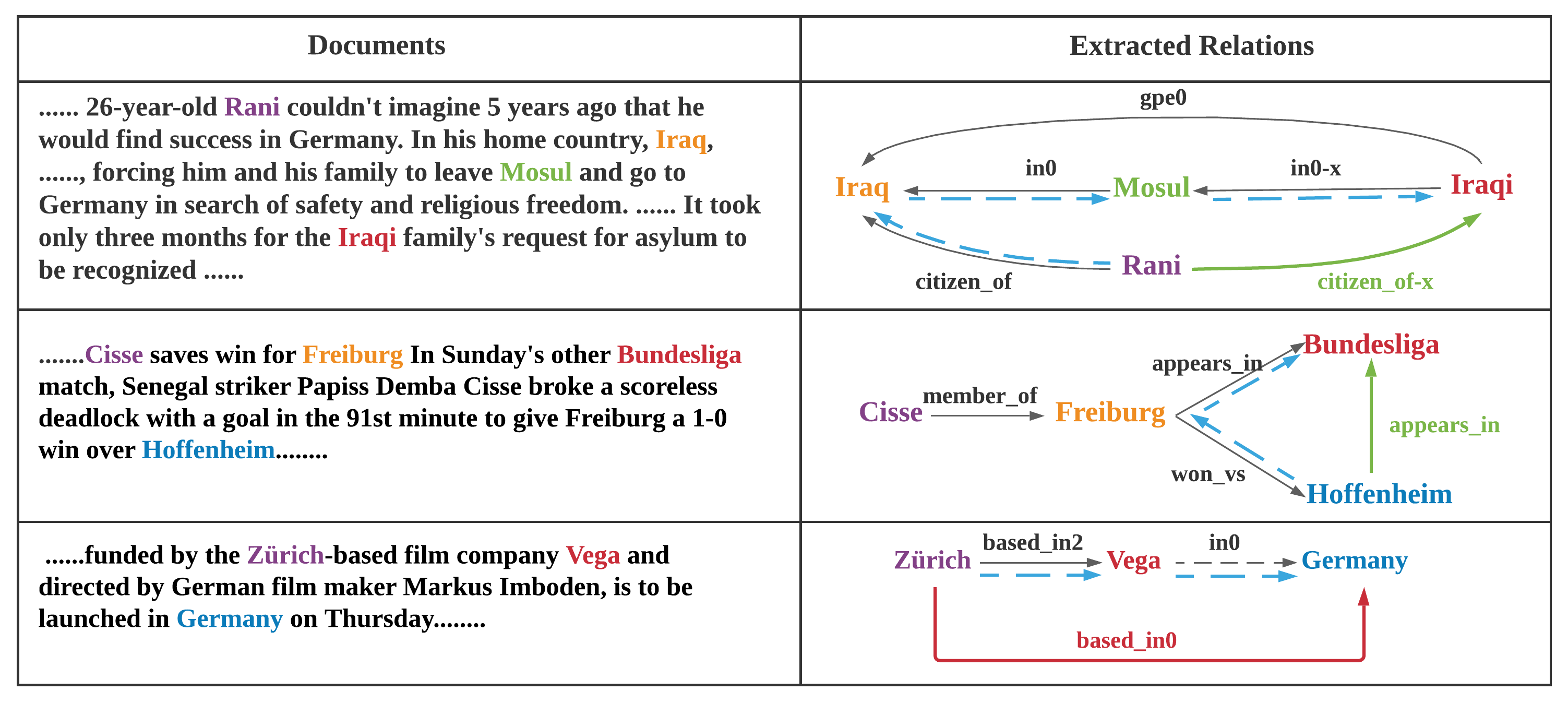}
    \caption{Inference cases of our \mymodel on DWIE by using ATLOP as the backbone model. The grey arrows are relations extracted by the backbone model, solid lines representing true relations while dashed lines representing false relations. The green arrows are new relations correctly extracted by logical reasoning. The blue arrows indicate the potential reasoning paths. We also demonstrate a negative case. In the third example, the red arrow represents a wrong relation extracted by reasoning over wrongly estimated atoms.}
    \label{fig:case-study}
\end{figure*}

\paragraph{Interpretability by Generating Rules}
Our \mymodel enjoys better interpretability with the generated latent rule set. After the EM optimization, we can sample from the rule generator for high-quality rules that may contribute to the final predictions. Besides the gold rules previously shown for evaluating \textbf{logic}, \mymodel mines more logic rules from the data, as shown in Table.~\ref{tab:induct}. These logic rules explicitly reveal the interactions among entities and relations in the same document as regular patterns. \mymodel is more transparent, exhibiting the latent rules by the rule generator.

\paragraph{Case Study}
Fig.~\ref{fig:case-study} shows a few inference cases of our \mymodel, including two positive examples and a negative one. As shown in the first two examples, \mymodel can complete the missing relations in the backbone model's outputs by utilizing logical rules. 
% The defects of representation-based approaches under specific circumstances can be remedied by the soft logic reasoning. 
The soft logical reasoning can remedy the defects of representation-based approaches under specific circumstances. 
However, the extra reasoning may also exacerbate errors by reasoning over wrongly estimated logic units. The third example shows such a case. The wrongly estimated atom \emph{in0}(Vega, Germany) leads to one more wrong relation extracted by reasoning. Fortunately, such errors in our \mymodel will be more controllable because of the transparency in the logical reasoning part.

%% file: sections/05conclusion.tex
In this paper, we proposed a probabilistic model \mymodel, which utilizes rules and conducts reasoning over the rules for document-level relation extraction. The logic rules are treated as latent variables. We utilize the EM algorithm to efficiently maximize the overall likelihood. By injecting rules to the relation extraction framework, our \mymodel explicitly models the long-range dependencies in docRE as interactions among relations and entities, thus enjoying better interpretability. Empirical results and analysis show that \mymodel outperforms strong baselines on overall performance, logical consistency, and capability for capturing long-range dependencies. 

%% file: sections/06appendix.tex
% \newpage
\appendix

% \twocolumn[
% \begin{@twocolumnfalse}
%     \section*{ \centering{ Supplementary Materials for \\ \emph{Learning Logic Rules for Document-Level Relation Extraction\\[30pt]}}}
% \end{@twocolumnfalse}
% ]

\section{Approximation of the True Posterior}
\label{sec:appendix:approx}

The exact posterior of the latent rule set $\bm{z}$ is difficult to be directly calculated because of the large space.
In this section, we provide the detailed derivation for the approximate posterior.
% \begin{equation*}
% \begin{aligned}
%     &\log p(\bm{z}|\bm{y},\bm{q},\mathcal{D}) \\
%     =& \log p_w(\bm{y}|\bm{q},\bm{z},\mathcal{D}) + \log p_\theta(\bm{z}|\bm{q}) + C \\
%     =& \log\frac{1}{1+e^{-\bm{y}\cdot\mathtt{score}(\mathbf{q}, \bm{z})}} + \sum_{z\in\bm{z}}\log p_\theta(z|\bm{q}) + C \\
%     \approx&\frac{1}{2} \bm{y}\cdot\mathtt{score}_w(\bm{q},\bm{z}) + \sum_{z\in\bm{z}}\log p_\theta(z|\bm{q}) + C \\
%     =&\sum_{\emph{rule}\in\bm{z}}(\frac{1}{2}\bm{y}\cdot(\frac{1}{N}(\phi_w(\bm{q}) + \phi_w(\bm{q},\emph{rule})\phi_w(\emph{rule})) \\ &+\log\text{AutoReg}_\theta(\emph{rule}|\bm{q})) + C
% \end{aligned}
% \end{equation*}
\begin{equation*}
\begin{aligned}
    &\log p(\bm{z}|\bm{y},\bm{q},\mathcal{D}) \\
    =& \log p_w(\bm{y}|\bm{q},\bm{z},\mathcal{D}) + \log p_\theta(\bm{z}|\bm{q}) + C \\
    =& \log\frac{1}{1+e^{-\bm{y}\cdot\mathtt{score}(\mathbf{q}, \bm{z})}} + \sum_{rule\in\bm{z}}\log\text{AutoReg}_\theta(rule|\bm{q}) + C \\
    \approx&\frac{1}{2} \bm{y}\cdot\mathtt{score}_w(\bm{q},\bm{z}) + \sum_{rule\in\bm{z}}\log \text{AutoReg}_\theta(rule|\bm{q}) + C \\
    =&\sum_{\emph{rule}\in\bm{z}}(\frac{1}{2}\bm{y}\cdot(\frac{1}{N}(\phi_w(\bm{q}) + \phi_w(\bm{q},\emph{rule})\phi_w(\emph{rule})) \\ &+\log\text{AutoReg}_\theta(\emph{rule}|\bm{q})) + C
\end{aligned}
\end{equation*}
The approximation is obtained by the following second-order Taylor expansion:
\begin{equation*}
    -\log(1 + e^{-x}) = -\log2 + \frac{x}{2} + O(x^2)
\end{equation*}
By such approximation, we can decompose the posterior to each rule in the latent rule set. We first define the score for each rule:

\begin{IEEEeqnarray*}{l}
H(\emph{rule}) = \log \mathrm{AutoReg}_\theta(\emph{rule} | \bm{q}) + \\
\frac{y^*}{2}\left(\frac{1}{N}\phi_w(q)+\phi_w(q,\emph{rule})\phi_w(\emph{rule})\right)
\end{IEEEeqnarray*}

Then, it's easy to obtain that the approximated posterior $q(\bm{z})$ and the prior $p_\theta$ are conjugate distributions.
\begin{equation*}
\begin{aligned}
    q(\bm{z}) \sim \text{Multi}(N, \frac{1}{Z}\exp(H(\emph{rule})))
\end{aligned}
\end{equation*}

where $Z$ is the normalization factor.

\begin{equation*}
\begin{aligned}
    \text{AutoReg}_\theta(rule|q) = \prod_{i=1}^l\text{AutoReg}_\theta(r_{i}|q, r_1, r_2, ..., r_{i-1})
\end{aligned}
\end{equation*}

\section{Implementation Details}
\label{sec:appendix:imp}

\paragraph{DWIE Dataset Preprocessing}
The original DWIE dataset \cite{zaporojets2020dwie} is designed for four sub-tasks in the information extraction, including named entity recognition, coreference resolution, relation extraction, and entity linking. In this paper, we focus on the document-level RE task. We only use the dataset for document-level relation extraction. The original dataset published 802 documents with 23130 entities in total, 702 for train and 100 for test. In our setting, we remove the entities without mentions in the context. After the cleaning, we have 700 documents for train and 99 documents for test. The training set is then randomly split into two parts: 602 documents for train and 98 for development. The statistics of the preprocessed dataset are shown in Table \ref{tab:dataset} of the main body.

\paragraph{Baselines}
We use their published open-source code to implement the baselines \cite{yao-etal-2019-docred, zeng-etal-2020-double, zhou2021atlop}, as well as the backbone models in our framework. The pre-trained language models used in GAIN and ATLOP follows the original paper \cite{zeng-etal-2020-double, zhou2021atlop}, using the pre-trained bert-base-uncased and bert-base-cased models respectively. The hyperparameters reserve the same as in their papers.